\documentclass[10pt]{article} 
\usepackage{rlc}

\usepackage{amssymb}            
\usepackage{mathtools}          
\usepackage{mathrsfs}           
\mathtoolsset{showonlyrefs}     
\usepackage{graphicx}           
\usepackage{subcaption}         
\usepackage[space]{grffile}     
\usepackage{url}                
\usepackage{booktabs}       
\usepackage{amsfonts}       
\usepackage{nicefrac}       
\usepackage{microtype}      
\usepackage{xcolor}         
\usepackage{amsmath,amsthm}
\usepackage{wrapfig}
\usepackage{graphicx}

\newcommand\blfootnote[1]{%
  \begingroup
  \renewcommand\thefootnote{}\footnote{#1}%
  \addtocounter{footnote}{-1}%
  \endgroup
}

\title{\centering What Makes It Difficult to Adapt a Reinforcement Learning Agent to New Tasks?}
\title{\centering Towards Adapting Reinforcement Learning Agents to New Tasks: Insights from Q-Values}

\begin{document}

\maketitle
\begin{center}
Ashwin Ramaswamy\footnote{~\href{mailto:ashwinsrama@gmail.com}{ashwinsrama@gmail.com}} and Ransalu Senanayake\\
Arizona State University
\end{center}

\blfootnote{Presented at the Robotics: Science and Systems (RSS) 2024 Workshop on Task Specification for General-Purpose Intelligent Robots}

\begin{abstract}
While contemporary reinforcement learning research and applications have embraced policy-gradient methods as the panacea of solving learning problems, value-based methods can still be useful in many domains, as long as we can wrangle with how to exploit them in a sample efficient way. In this paper, we explore the chaotic nature of DQNs in reinforcement learning, while understanding how the information that they retain when trained can be repurposed for adapting a model to different tasks. We start by designing a simple experiment in which we are able to observe the Q-values for each state and action in an environment. Then we train in eight different ways to explore how these training algorithms affect the way that accurate Q-values are learned (or not learned). We tested the adaptability of each trained model when retrained to accomplish a slightly modified task. We then scaled our setup to test the larger problem of an autonomous vehicle at an unprotected intersection. We observed that the model is able to adapt to new tasks quicker when the base model's Q-value estimates are closer to the true Q-values. The results provide some insights and guidelines into what algorithms are useful for sample efficient task adaptation.
\end{abstract}

\section{Introduction}
\label{sec:submission}

Reinforcement Learning (RL) has been demonstrated as a promising framework for various downstream tasks such as Large Language Models (LLMs) \citep{ouyang2022traininglanguagemodelsfollow}, robotics \citep{Ibarz_2021}, and finding failure modes in generative models \citep{Sagar2024icml}, due to the malleability \citep{popov2017dataefficientdeepreinforcementlearning} of an agent's policy, $\pi_\theta$, through a reward function, $R(s, a)$. The full process of RL usually involves collecting data from the environment, utilizing a parameterized policy to generate an estimate of the reward function, enacting the policy in the environment, and then updating the policy with respect to the outcome of the enacted episode. Contemporary research in RL explores ways to learn the optimal policy for a specific task by limiting interaction with the environment (sample efficiency) \citep{gou2019dqnmodelbasedexplorationefficient} and ensuring that the learned policy does not significantly deviate from the expected behavior (stability) \citep{ross2011reductionimitationlearningstructured}. While Deep Q-Networks (DQNs) have shown success in some areas of reinforcement learning, they have been known to be incredibly sample inefficient \citep{hessel2017rainbowcombiningimprovementsdeep} and sensitive to the task and experimental setup. Changing the hyperparameters, architecture, reward function, and even the random seeding \citep{coad2021blog} of a DQN model by a small amount can have dramatic consequences on how well an agent learns. 
In this regard, although \emph{policy gradient methods} are currently commonplace in research settings \citep{karpathy2016blog}, DQNs serve as a valuable educational tool to explore the limits and new ideas of integrating deep neural networks with Reinforcement Learning. For example, DQNs have been shown to work in game-based settings such as Atari games \citep{mnih2013atari}, which has jump-started the feasibility of RL as a catalyst for general-purpose AI. 

To complicate the implementation of RL, a policy that is learned for a specific task may become ineffective if the environment's constraints change, new states and actions emerge, or the desired task itself is modified, as the original reward function may no longer guide the agent to successfully perform the intended behavior for these new conditions \citep{Toro_Icarte_2022}. Any such change often requires the data collection process to be restarted from scratch \citep{Noel2024}. \emph{Transfer Learning} attempts to tackle this issue by utilizing the learned data for multiple tasks. It has shown to work well for supervised learning tasks like image classification and sentiment analysis \citep{zhuang2020comprehensivesurveytransferlearning}. This is because fine-tuning the model for the new task can be as simple as involving backpropagation with respect to the new dataset to generalize. However, Transfer Learning is difficult to reproduce for downstream tasks in the RL domain because RL is a controls problem, where the goal is to estimate a policy that generates a full trajectory, not just a one-step estimation of labels. The work in \citep{zhang2024extractefficientpolicylearning} enables sample efficient transfer learning through skill extraction, but requires the "initial dataset from environments similar to the target environments for learning skills from". Additionally, it focuses on learning specific skills and training an agent on those skills for a new task. In cases where the skill is still being developed or is highly abstracted for a task \citep{jaquier2024transferlearningroboticsupcoming}, it is unclear whether this implementation have the same effectiveness. To the best of our knowledge, the sample efficiency of fine-tuning robot models for the transferred task has not been explored in depth for \textbf{agents using DQNs} as the underlying model architecture. Consequently, \emph{Meta-Learning} has been shown to be successful to generalize to new RL tasks with few-shot success \citep{finn2017maml}. However, there are some feasibility issues---consider the robotics domain. In order to train a general model to be sensitive to changes in a task, the robot would need to train on different datasets of sampled trajectories from a distribution of tasks. This can be extremely sample inefficient and impractical to implement for robots that interact with the physical world, where data collection for each task in the training set would be time-consuming \citep{bousmalis2017usingsimulationdomainadaptation}. Therefore, the benefits of Meta-Learning are less likely to emerge in cases where a model has been trained specifically for one downstream task, rather than on a distribution of tasks. Retroactive generalization for the model is not possible in the meta-learning setup. Since the problem of task adaptability is agnostic to the underlying RL method---DQN, policy gradient, inverse optimal control, etc.,---this paper explores task adaptation on a \textbf{DQN} when trained to accomplish a \textbf{single task}.

\subsection{Preliminaries}
Deep Q-Learning is a method of deep reinforcement learning that uses neural networks to approximate Q-values at a given state. A Q-value is the expected return when taking an action at a state, which is used as a greedy search heuristic for determining the optimal policy for an agent. The general idea in Q-learning is that by sampling states, actions, proximal states, and rewards in an environment, an agent can learn the intermediary associations between actions and rewards that will allow an agent to take the most optimal action at a given state, by maximizing for the cumulative earned reward \citep{watkins1992q}. The Q-value for a given state and action can be computed as: 
$$Q(s_{t}, a_{t}) = r(s_t) + \gamma \cdot \max_{a \in A}Q(s_{t+1}, a_{t+1}) \eqno{(1)}$$ 
where $s_t, a_t$ are the current state-action values and $s_{t+1}, a_{t+1}$ are the next state-action values for a discount factor $\gamma$. 

However, with an intractably large state space, it is infeasible to store the Q-value of each possible action. Therefore, a deep neural network commonly referred to as a Deep Q-Network (DQN) is used to approximate the Q-values for each action by taking the features of a state as input, and returning the estimated Q-value of for each action as the output. The logic behind the Deep Q-Learning implementation is as follows:
\begin{enumerate}
    \item Given a state input, the model predicts an action output referred to as ``pred.'' The output represents a tensor of the initial estimates of Q-values for each of the actions at a state.
    \item The output tensor is cloned and referred to as ``target.'' This tensor will be updated to match a closer estimate of the true Q-values, calculated in the subsequent step.
    \item For each step in the trajectory (a state-action taken), from the available and trained data, we define $Q_{new}$ as the reward earned taking that step plus the discounted maximum possible Q-value that the model predicts for the next state.
    \item The maximum Q-value in “target” is replaced with this new Q-value.
    \item MSE Loss is calculated between the ``pred'' and ``target'' tensors. The idea is to update the parameters of the model so that it is able to accurately predict the tensor for ``target.'' 
\end{enumerate}

We expect that with enough samples, the model will be accurately able to predict the correct Q-values for all states and actions, because the future Q-values at states eventually inform the Q-values at earlier states on successive updates.

\subsection{Task Adaptation}

When an agent is has been trained with a DQN to accomplish a task, its learned Q-values have been updated to estimate optimal actions for just that task. Therefore, if the agent needs to accomplish a different task, it will need to retrain its DQN from scratch to learn the correct Q-values to accomplish the new task. This requires sampling more states, actions, and next states to learn the reward associations. In the interest of reusing learned associations from the original task, we investigate a simple idea: retraining the agent to perform new task directly on the neural network trained for the original task. Inspired by transfer learning, the idea is that when training on the original task, the model learns an optimal policy and gains some inherent representation of the environmental dynamics that help it solve the task \citep{neyshabur2021transferredtransferlearning}. When defining a new task we consider the new task as a small modification of the feature space of the reward function.

\noindent {\bf Hypothesis:} If the new task is not significantly different from the original task, then we hypothesize that the weights learned for the new task are within some difference of the original neural network such that fine-tuning the original DQN is more sample efficient than training from scratch. 

To this extent, we explore how different training algorithms to generate trajectories for DQNs enable fast learning. We conduct experiments using on-policy training, random exploration, expert demonstrations, and then compare these to the model's performance when trained with supervised learning instead of any deep reinforcement learning method.

\section{Experiment 1: A Grid}

\subsection{Experimental Setup}
The goal of reinforcement learning is to train an agent to accomplish a task by using rewards as an auxiliary value that influences actions made by the agent towards achieving the goal task. It operates on the delicate balance between exploration and exploitation. In order to understand how this process allows the agent to learn, we have crafted a simple experiment to observe the way an agent learns Q-values and how it affects the way it learns to accomplish the task. We ran each training run of the experiment multiple times, so while there is some stochasticity due to random seeding, the results are generally consistent and representative of the expected results. Consider a 3-by-3 grid that has a green cell, red cell, a brown cell, and 6 other white cells. The goal of the agent is to traverse the grid and arrive at the green cell without going out of bounds or hitting the red cell, both of which would terminate the game. It must try to avoid the brown cell, which represents an obstacle and merely provides a penalty. The agent can take the actions ``up,'' ``down,'' ``left,'' or ``right'' at any non-terminal (white) cell and is not penalized in the first time step. This experimental setup uses a non-stochastic policy and the reward function specified in Fig.~1.

\begin{figure}[htbp]
    \centering
    \begin{minipage}[b]{0.45\textwidth}
        \centering
        \begin{tabular}{ll}
            \toprule
            Agent is: & Reward \\
            \midrule
            at green cell & +20, terminal \\
            at red cell (1, 2) & -20, terminal  \\
            at brown cell (1, 0) & -5, non-terminal  \\
            out of bounds & -100, terminal  \\
            frozen ($>$ 10 steps) & -100, terminal  \\
            at white cell & -1, non-terminal \\& if steps < 10  \\
            \bottomrule
        \end{tabular}
        \caption{Experiment 1: The Q-values for each state. The green and red squares represent the ``good'' and ``poor'' terminal states, respectively. Triangles that are highlighted in light blue represent the optimal action at each state, corresponding to the maximum Q-value at each state. The reward function used is described in the table above.}
    \end{minipage}
    \hfill
    \begin{minipage}[b]{0.45\textwidth}
        \label{fig:exp1}
        \centering
        \includegraphics[width=\textwidth]{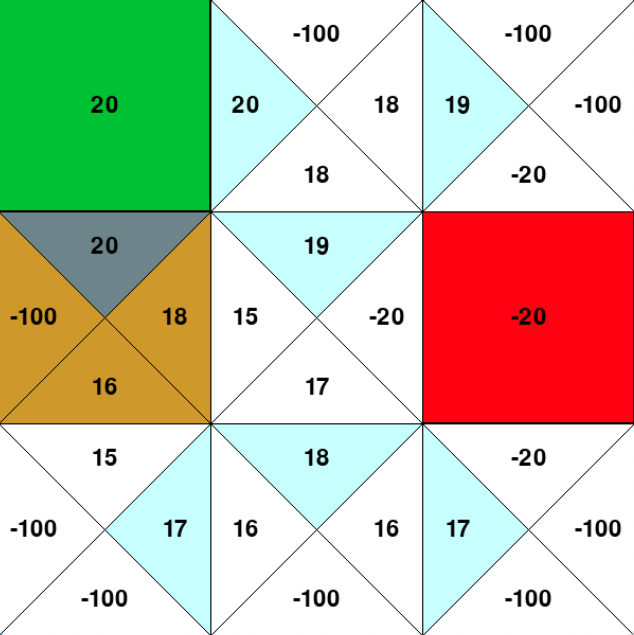}
    \end{minipage}
\end{figure}

Applying a dynamic programming formulation for Eq. 1, we were able to hand calculate the Q-value for each action at each state, as depicted in Fig. 1. The highlighted blue triangles in each state correspond to the action of the optimal policy in that state. Regardless of the methodology used to train the DQN, we expect that the agent eventually learns the optimal policy to reach the green cell. Our model is parameterized as a three-layer neural network that takes the state as input (y-pos, x-pos), connects to a 512 neuron hidden layer activated by ReLU, and has 4 neurons in its output layer, corresponding to the actions for up, down, left, and right. The learning rate is 0.0001 and $\gamma$ is 1, as there is already a -1 living penalty after the first time step. Training is conducted for 20,000 episodes. 

In order to gauge the success of the model while it is being trained, we implement a ``test while train'' approach, in which we utilize an ``accuracy'' metric, defined as the percentage of episodes where the agent successfully accomplishes the task out of 250 simulated roll-outs of the currently training model after every 10 episodes of training. In this experiment, a success is any trajectory that includes the green cell (0, 0) and does not include the brown cell (1, 0). Due to the terminality setup, by definition any trajectory that contains the red cell or goes out of bounds will not also contain the green cell. For the purpose of experimentation, in this setup the expert’s actions are also optimal (i.e. ``expert policy'' and ``optimal policy'' are used interchangeably.) This is a strict assumption to make, as it may not be the case for autonomous driving, or other dynamic environments. If this condition is not true in a setting, then the agent's learning will converge to the expert demonstrator’s behavior rather than the optimal behavior.

Using this experimental setup, we make the agent learn using one or a combination of training algorithms: 1) On-policy - the agent takes the action corresponding to the maximum Q-value output by the neural network, 2) Random Exploration - the agent takes a random action at a state, and 3) Expert Demonstration - the agent queries an expert model that always takes the optimal action at any state, and then takes that action. As a control, we also train a model using supervised learning where the agent can query an expert that knows the optimal Q-value and action at every state.

\subsection{Results for the Original Task}
In order to evaluate the efficacy of each training algorithm, we employ an MSE criterion to quantify how well the trained model is able to estimate the true Q-values for each state and action. After training for initial task success, we see that after 20,000 episodes the agent is able to achieve 100\% task accuracy only when alternating between on-policy evaluation and querying the expert. As depicted in Table 1, this yields a very low MSE across the truly optimal trajectory, but a high MSE elsewhere. This makes sense because the agent never does random exploration, so it is more likely to follow converge to the expert demonstrator's behavior. Consequently, we see that training the agent by alternating between random exploration and on-policy evaluation yields a relatively low MSE across all Q-states but a moderately high MSE across the truly optimal trajectory. This also makes sense because random exploration forces the agent to visit states and take actions that it would never be incentivized to do under an on-policy evaluation. The on-policy evaluation on the other hand is able to adjust Q-values that are more aligned with the model's predictions. Supervised learning does not involve Q-values, and just trains a model to match the predicted action with the true action of the expert, and therefore is quick to yield a 100\% task accuracy. Random exploration allows the agent to update its Q-values for rarely seen state-action pairs (the agent has no incentive to take some action or visit some state), yielding significantly slower convergence to 100\% task accuracy but more accurate Q-values when combined with other policies. As the number of training episodes tends towards infinity, the Q-value accuracy grained by random exploration will reach optimal values.

In all of the experiments, after running for longer trials, they are all able to reach a 100\% task accuracy rate with random exploration. The results beget the question: why does it matter if all of the Q-values are accurate if the agent learns to successfully accomplish the task? What if the agent is just able to accomplish the task but has poor Q-value estimates? If the agent learns accurate Q-values along the optimal trajectory, but poor Q-values elsewhere, then due to compounding errors, the agent may take sub-optimal actions when it is no longer on the optimal trajectory \citep{luo2024rlif}.

\begin{table}[t]
\caption{Results for the original task in experiment 1. ACC indicates Task Accuracy after 20,000 Training Episodes, MSE* indicates MSE across Optimal Q-states, and MSEq indicates MSE across all Q-states.}
\begin{tabular}{lrrr}
\toprule
\textbf{Algorithm}                                                                              & \textbf{\begin{tabular}[c]{@{}c@{}}ACC\end{tabular}} & \textbf{\begin{tabular}[c]{@{}c@{}}MSE*\end{tabular}} & \textbf{\begin{tabular}[c]{@{}c@{}}MSEq\end{tabular}} \\ \midrule
1. On-Policy                                                                                       & 0.86                                                                                         & 38.27                                                                           & 1047.29                                                                    \\ 
2. Random Explore                                                                                  & 0.86                                                                                         & 1325.97                                                                         & 992.33                                                                     \\ 
3. Expert Demos                                                                                    & 0.14                                                                                         & 19.74                                                                           & 4808.63                                                                    \\ 
\begin{tabular}[c]{@{}c@{}}4. Alternate (Random Explore + On-Policy)\end{tabular}               & 0.43                                                                                         & 400.01                                                                          & \textbf{594.44}                                                            \\ 
\begin{tabular}[c]{@{}c@{}}5. Alternate (Random Explore  + Expert Demos)\end{tabular}            & 0.86                                                                                         & 438.13                                                                          & 768.79                                                                     \\ 
\begin{tabular}[c]{@{}c@{}}6. Alternate (On-Policy + Expert Demos)\end{tabular}                 & \textbf{1.00}                                                                                  & \textbf{0.0019}                                                                 & 1100.96                                                                    \\ 
\begin{tabular}[c]{@{}c@{}}7. Alternate (Random Explore + On-Policy + Expert Demos)\end{tabular} & 0.86                                                                                         & 241.59                                                                          & 693.63                                                                     \\ 
8. Supervised Learning                                                                             & \textbf{1.0}                                                                                  & 5791.49                                                                         & 4973.93                                                                    \\ \bottomrule
\end{tabular}
\end{table}

\subsection{Results for the Adapted Task}
With sufficient training through deep Q-learning or supervised learning, the agent is able to successfully accomplish the original task. Therefore, we use these trained models as base models to evaluate how well the agent is able to adapt to a new task: \emph{having the agent to get to the red state and avoid the green cell, effectively swapping the functionalities of the red and green cells from the original task}. The only change in the reward function is to swap the penalties for the agent when existing at the green and red states. The agent will be loaded with one of the base models and will use deep Q-learning to train on top of the base model, effectively treating it as a DQN. 

Table 2 shows the results of retraining the agent using the supervised learning model as a base model, where effectively all optimal actions for the original task are one-hot encoded. It also displays the results when using the optimally trained DQN as the base model. Both base models were retrained with all of the learning algorithms used in the original task. The results are surprising because they show that the agent can quickly converge to accurate Q-values for the new task using Deep Q-learning even if it was trained with supervised learning for the original task, where the estimated Q-values are significantly different. Similarly to the results from Table 1, training the neural network with an alternation of expert demonstrations and on-policy action selection yields a very quick convergence for 100\% task success of the new task and the lowest MSE across optimal states with both the supervised learning and optimal DQN base models. Training with solely on-policy selection or solely expert demonstrations does not result in the agent converging to 100\% task accuracy within 20,000 training episodes regardless of the base model. 

We also observe that involving random exploration in the retraining stage consistently reduces the MSE across all Q-states regardless of the base model, relative to agents that train without random exploration, consistent with the findings in \citep{tijsma2016exploration}. Finally, agents that train with supervised learning are able to settle at 100\% new task accuracy in the least number of samples, regardless of the base model, but have the highest MSE across all optimal and Q-states, suggesting that the agent would perform poorly if out of distribution. Consequently, when comparing the MSE results for each retraining algorithm between base models, agents that train with the optimal DQN base model have significantly lower MSEs across optimal Q-states and all Q-states than agents trained on the supervised learning base model. The number of episodes to settle at 100\% accuracy are generally similar, so the main advantage of training on the DQN is closer to correct Q-values for the new task.

\begin{table}[t]
\caption{Results for the adapted task in experiment 1. EPI indicates number of episodes to settle at 100\% Accuracy, MSE* indicates MSE across Optimal Q-states, and MSEq indicates MSE across all Q-states.}
\begin{tabular}{lrrr}
\toprule
\textbf{\begin{tabular}[c]{@{}c@{}}\end{tabular}}  \textbf{Retraining Algorithm}                                                                   & \textbf{\begin{tabular}[c]{@{}c@{}}EPI\end{tabular}} & \textbf{\begin{tabular}[c]{@{}c@{}}MSE*\end{tabular}} & \textbf{\begin{tabular}[c]{@{}c@{}}MSEq\end{tabular}} \\ \midrule
{\bf Pre-trained model: one-hot encoded} \\
On-Policy                                                                                       & \begin{tabular}[c]{@{}c@{}}settles at 23\%\end{tabular}                       & 7322.61                                                                         & 5185.93                                                                    \\ 
Random Explore                                                                                  & 7220                                                                              & 304.08                                                                          & 259.00                                                                     \\ 
Expert Demos                                                                                    & \begin{tabular}[c]{@{}c@{}}settles at 22\%\end{tabular}                       & 1396.22                                                                         & 4232.36                                                                    \\ 
\begin{tabular}[c]{@{}c@{}}Alternate (Random Explore + On-Policy)\end{tabular}               & 9110                                                                              & 226.09                                                                          & 327.16                                                                     \\ 
\begin{tabular}[c]{@{}c@{}}Alternate (Random Explore + Expert Demos)\end{tabular}            & 5680                                                                              & 101.39                                                                          & 211.06                                                                     \\ 
\begin{tabular}[c]{@{}c@{}}Alternate (On-Policy + Expert Demos)\end{tabular}                 & 3320                                                                              & \textbf{0.0036}                                                                 & 1823.15                                                                    \\ 
\begin{tabular}[c]{@{}c@{}}Alternate (Random Explore + On-Policy + Expert Demos)\end{tabular} & does not settle                                                                   & 1077.19                                                                         & 3704.49                                                                    \\ 
Supervised Learning                                                                             & 1300                                                                              & 2937.29                                                                         & 5662.49                                                                    \\ \midrule
{\bf Pre-trained model: optimal Q-value} \\
 On-Policy                                                                                       & \begin{tabular}[c]{@{}c@{}}settles at 0\%\end{tabular}                        & 4311.63                                                                         & 2495.82                                                                    \\ 
 Random Explore                                                                                  & 9390                                                                              & 12.72                                                                           & 13.28                                                                      \\ 
Expert Demos                                                                                    & \begin{tabular}[c]{@{}c@{}}settles at 0\%\end{tabular}                        & 370.84                                                                          & 755.29                                                                     \\ 
 \begin{tabular}[c]{@{}c@{}}Alternate (Random Explore + On-Policy)\end{tabular}               & 4670                                                                              & 61.88                                                                           & 74.32                                                                      \\ 
\begin{tabular}[c]{@{}c@{}}Alternate (Random Explore + Expert Demos)\end{tabular}            & 11780                                                                             & 2.11                                                                            & \textbf{11.02}                                                             \\ 
\begin{tabular}[c]{@{}c@{}}Alternate (On-Policy + Expert Demos)\end{tabular}                  & 1590                                                                              & 1.77                                                                            & 527.52                                                                     \\ 
\begin{tabular}[c]{@{}c@{}}Alternate (Random Explore + On-Policy + Expert Demos)\end{tabular} & 3890                                                                              & 463.58                                                                          & 205.75                                                                     \\ 
Supervised Learning                                                                             & \textbf{60}                                                                       & 3665.87                                                                         & 3233.46                                                                    \\ \bottomrule

\end{tabular}
\end{table}

\section{Experiment 2: Autonomous Intersection Crossing}
\label{sec:margins}

\subsection{Experimental Setup}
Since the 3$\times$3 grid world experiments above are rudimentary and rather arbitrary for a task adaptability study, we tested our hypothesis on a scaled up and practical task: autonomous intersection crossing. The initial task is for a car (the ego car) in one lane to safely cross an intersection against oncoming traffic (ado cars) that has the right of way in a perpendicular lane, but only if the gap between two ado cars that the ego car passes through is at least 80 pixels wide. The environment is represented as two perpendicular one-way lanes. The cars are represented as circles with an x and y coordinate corresponding to their center. 

The DQN has 16 inputs, a hidden layer of 1024 neurons, and an output layer of 2 neurons, corresponding to the two possible actions; stop and go at a fixed velocity. The first two layers are activated by ReLU. Each opposing car is spawned randomly, resulting in a stochastic and dynamic environment unlike the grid world setting. The view of the ego vehicle is set up in a way that it can see up to 10 ado cars. Therefore, the features of each state include, the y coordinates of the ado cars, the x and y coordinates of the ego car, the floating point distance from the ego car to the other side of the intersection, a Boolean value of whether the ego car is in the intersection, a Boolean value of whether there exists a gap of 80 between two ado cars in front of the intersection, and a Boolean value of whether there exists a gap of 120 between two ado cars in front of the intersection. 

\begin{figure}[htbp]
    \centering
    \begin{minipage}[b]{0.45\textwidth}
        \centering
        \begin{tabular}{ll}
            \toprule
            Car status: & Reward \\
            \midrule
            crossed intersection \\ safely (gap $\geq$ 80) & +3000, terminal \\
            crossed intersection \\ unsafely (gap $<$ 80) & -9500, terminal \\
            collision & -9500, terminal  \\
            frozen & -20000, terminal  \\
            forward progress & -1, non-terminal  \\
            no forward progress & -6, non-terminal  \\
            \bottomrule
        \end{tabular}
        \caption{Experiment 2: The simulated environment. Cars are represented as circles. The ego vehicle is the red circle travelling horizontally to the left of the road. The ado vehicles are the other colored circles travelling vertically downwards in the perpendicular road. The reward function
used is described in the table above.}
    \end{minipage}
    \hfill
    \begin{minipage}[b]{0.45\textwidth}
        \label{fig:exp1}
        \centering
        \includegraphics[width=\textwidth]{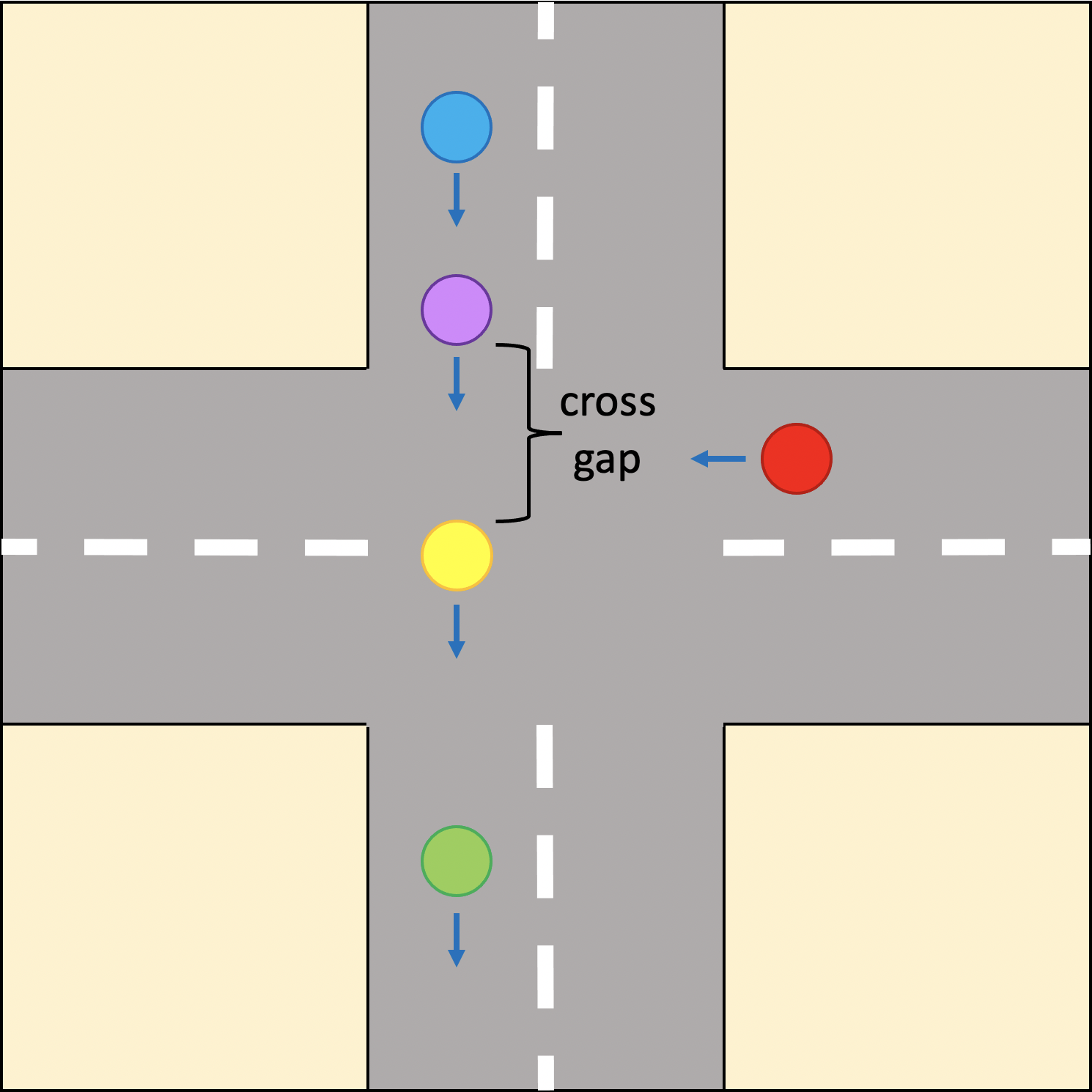}
    \end{minipage}
\end{figure}

\subsection{Results for the Original Task}
\begin{figure}[h!]
    \centering
    \includegraphics[width=1
\linewidth]{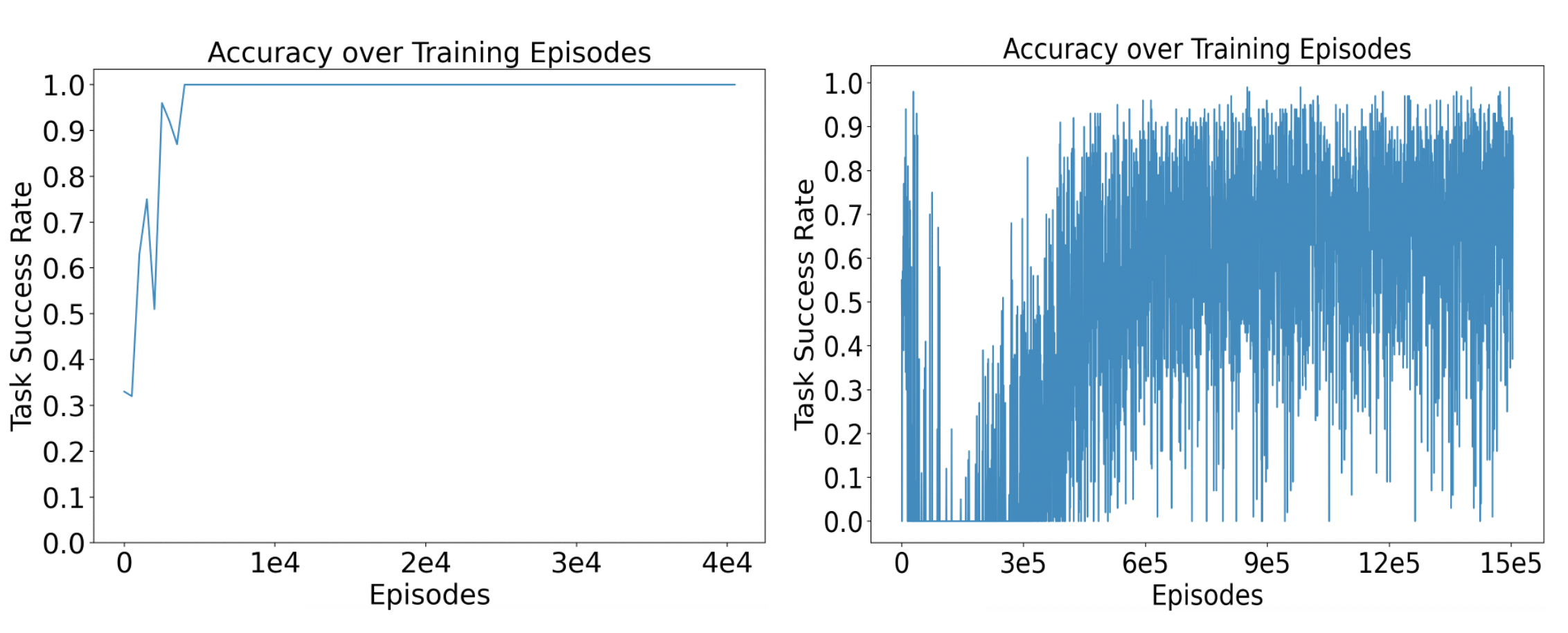}
    \caption{The graph on the left plots the \underline{task accuracy as the model trains with supervised learning}. The graph on the right plots the \underline{task accuracy as the model trains with deep Q-learning}.}
        \label{exp2:orig}
\end{figure}

For the original task, we trained a DQN using standard Deep Q-learning, which took 1.5 million episodes to converge (with high variance) to a task accuracy of greater than 75\%. The agent trained by sampling random actions and expert actions in alternating episodes in order to gain some off-policy accuracy. We separately trained a supervised learning model that converged to a 100\% task accuracy within less than 10,000 episodes (Fig.~\ref{exp2:orig}). Unlike the DQN model, this model was trained with access to the expert’s optimal actions at every possible state, making it the equivalent of standard behavioral cloning. Clearly, we see that with significantly less episodes, the model trained with supervised learning is able to cross the intersection safely, even in the presence of stochastically spawned ado cars. Consequently, the model trained with deep Q-learning takes millions of episodes to learn to safely cross the intersection even 75\% of the time.  

\subsection{Results for the Adapted Task}
The new task for the agent to adapt to is to cross the intersection only if the gap between the ado cars, while crossing the intersection, is at least 120 pixels wide. We chose this target task because it is not too different from the original task; merely a stylistic change in the nature of crossing predicated on the positions of the ado cars, which is observable through the state features. In practice this can be thought as answering the question: do we need to retrain an autonomous vehicle when the National Highway Traffic Safety Administration makes slight changes to safety standards?

We are able to use expert demonstrations in retraining because we have access to a sample of expert trajectories to show the model what the new task looks like. However, we must use deep Q-learning and not supervised learning because we cannot assume that we will have the expert’s action at every state. Because the agent just has access to a sample of expert demonstrations, we want to explore how long it will take for the agent to learn the new task using retraining, where the new reward function assigns positive reward to crossing only if the gap size is greater than 120, and is otherwise treated as crashing. Retraining on the supervised learning base model and DQN base model are both done with deep Q-learning. When adapting the model to the new task by retraining on the pretrained neural network as a starting point, we observe that it almost instantly settles to above 90\% task accuracy, which is significantly less than the 1.5 million episodes for initial training. Consequently, when retraining on the supervised learning based model, there is significantly more variance in the accuracy, and the agent does not seem to fully converge to 100\% task accuracy within the same number of training episodes. We find that the results are generally consistent across the same training parameters with the 3$\times$3 grid world experiments. One caveat is that in the grid world experiments the base models that the retraining was performed upon had accurately the optimal Q-values for the initial task and the one-hot encoded values for the DQN and SL models respectively, for all possible states and actions. This is infeasible for the DQN to learn in the initial task in a high dimensional space with many states, and hence, an approximation had to suffice where many off of optimal trajectory Q-values would be drastically incorrect. 

 \begin{figure}[h!]
    \centering
    \includegraphics[width=1
\linewidth]{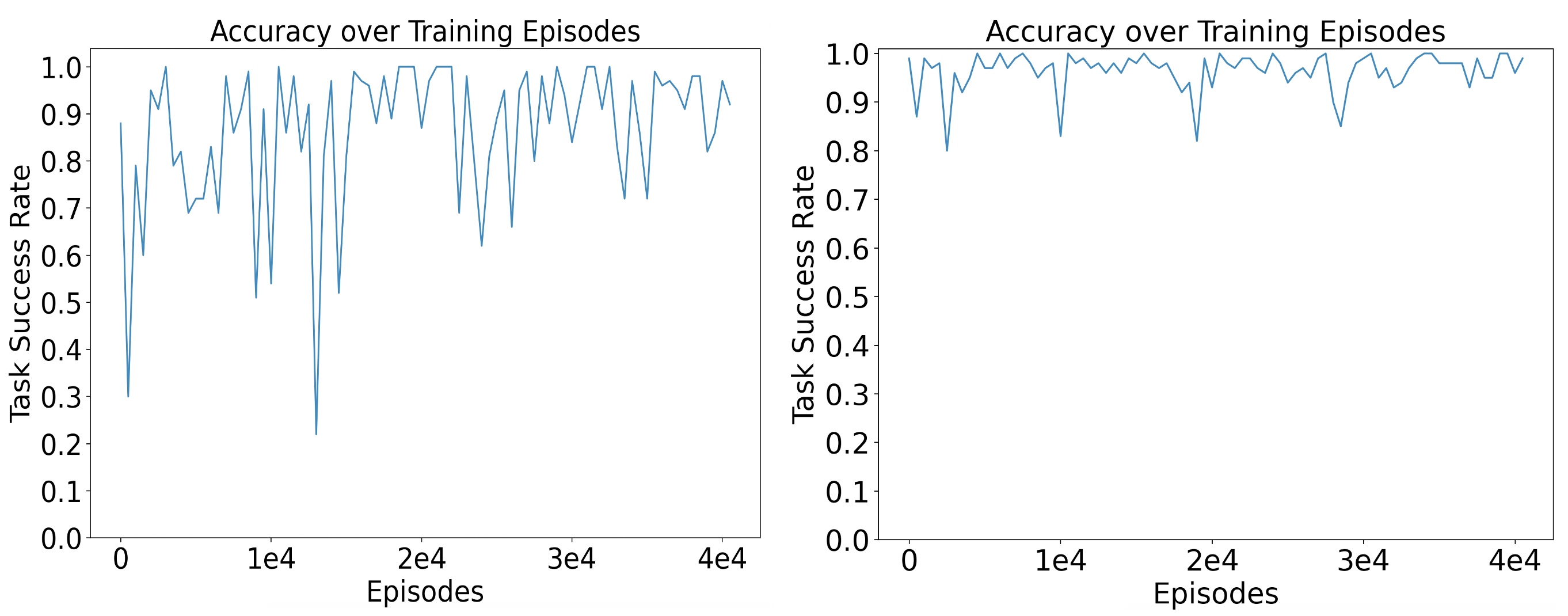}
    \caption{The graph on the left plots the accuracy of the model as it learns to adapt to the new task using the \underline{pre-trained supervised learning model as its base model}. The graph on the right plots the accuracy of the model as it learns to adapt to the new task using the \underline{pre-trained DQN model as its base model}.}
\end{figure}

\section{Conclusions}
\label{sec:margins}
We explored the trade-off between having the model estimate accurate Q-values and achieving sample-efficient task success. Furthermore, we learned the considerations for adapting a model to accomplish a different task both by retraining a supervised learning model and a DQN. When using a base model that has Q-values closer to the accurate Q-values, the adapted model is more likely to converge at 100\% accuracy as opposed to bouncing around, and achieve this convergence much quicker than when adapting from the supervised learning base model that does not estimate accurate Q-values. Furthermore, although random exploration increases the number of training samples that an agent will experience, it always helps in learning Q-values more accurately, and facilitates convergence when combined with methods that simply learn on the optimal path. We believe that this analysis will have diverse applications, including legacy systems where models have been trained for one purpose and used for years before needing to be updated or retrofitted by leveraging learned information.

\bibliography{main}
\bibliographystyle{rlc}

\end{document}